\def\year{2022}\relax
\documentclass[letterpaper]{article} 
\usepackage{aaai22}  
\usepackage{times}  
\usepackage{helvet}  
\usepackage{courier}  
\usepackage[hyphens]{url}  
\usepackage{graphicx} 
\usepackage{natbib}  
\usepackage{caption} 
\DeclareCaptionStyle{ruled}{labelfont=normalfont,labelsep=colon,strut=off} 
\usepackage{algpseudocode}
\usepackage{algorithm}
\usepackage{amsmath}
\usepackage{amsfonts}

\DeclareMathOperator*{\argmin}{argmin}

\title{Combining Reinforcement Learning and Optimal Transport for the Traveling Salesman Problem}
\author {
    Yong Liang Goh,\textsuperscript{\rm 1}
    Wee Sun Lee,\textsuperscript{\rm 1}
    Xavier Bresson,\textsuperscript{\rm 1}
    Thomas Laurent,\textsuperscript{\rm 2}
    Nicholas Lim\textsuperscript{\rm 3}
}
\affiliations {
    \textsuperscript{\rm 1} National University of Singapore\\
    \textsuperscript{\rm 2} Loyola Marymount University\\
    \textsuperscript{\rm 3} Grabtaxi Holdings Pte Ltd\\
    gyl@u.nus.edu, leews@comp.nus.edu.sg, xavier@nus.edu.sg, tlaurent@lmu.edu, nic.lim@grab.com
}

\begin{document}
\maketitle



\begin{abstract}
    The traveling salesman problem is a fundamental combinatorial optimization problem with strong exact algorithms. However, as problems scale up, these exact algorithms fail to provide a solution in a reasonable time. To resolve this, current works look at utilizing deep learning to construct reasonable solutions. Such efforts have been very successful, but tend to be slow and compute intensive. This paper exemplifies the integration of entropic regularized optimal transport techniques as a layer in a deep reinforcement learning network. We show that we can construct a model capable of learning without supervision and inferences significantly faster than current autoregressive approaches. We also empirically evaluate the benefits of including optimal transport algorithms within deep learning models to enforce assignment constraints during end-to-end training.
\end{abstract}

\section{Introduction}
\label{sec:intro}


The traveling salesman problem (TSP) is a well-studied NP-hard combinatorial problem. The problem asks: given a set of cities and the distances between each pair, what is the shortest possible path such that a salesman (or agent) visits each city exactly once and returns to the starting point? Variants of this problem exist in many applications such as vehicular routing \cite{robust1990implementing}, warehouse management \cite{zunic2017design}. Traditionally, the TSP has been tackled using two classes of algorithms; exact algorithms and approximate algorithms \cite{anbuudayasankar2014survey}. Exact algorithms aim at finding optimal solutions but are often intractable as the scale of the problem grows. Approximate algorithms sacrifice optimal solutions for computational speed, often providing an upper bound to the worst-case scenarios.

In this work, we focus on integrating deep learning to solve such combinatorial problems. Deep learning models have made significant strides in computer vision and natural language processing. Essentially, they serve as powerful feature extractors which are learnt directly from data. In the space of combinatorial problems, we ask: can deep learning replace hand-crafted heuristics by observations in data? This question is non-trivial due to the following:

\begin{itemize}
    \item Deep learning lies in continuous spaces, allowing for gradient-based methods. However, solutions to combinatorial problems are often discrete.
    \item Solutions for combinatorial problems are often non-unique, there can be more than one solution.
    \item Current deep learning solutions for such problems are memory and compute-intensive due to the need for a learnable decoder module.
    \item Exact solvers do not scale well with problem size, limiting the availability of labels for supervised learning.
\end{itemize}

In this work, we marry the previous approaches in supervised learning and reinforcement learning. Our contributions are as follows:
\begin{enumerate}
    \item Our model learns an edge probability via reinforcement learning, alleviating the need for labelled solutions.
    \item We enable much faster solutions by removing the need for a learnable decoder. 
    \item We improve the quality of the model by enforcing assignment constraints during end-to-end training of our model.
\end{enumerate}

While our model no longer takes advantage of the inductive bias of sequential encoding, we achieve improved speeds in both training and inference by producing the edge probabilities directly. Additionally, we showcase the positive impact of a differential optimal transport algorithm in \cite{cuturi2013sinkhorn} for the TSP. Our work also differs from previous attempts to integrate optimal transport such as \cite{emami2018learning} in two folds: (1) we leverage the transformer architecture that has superseded RNN-like approaches, (2) instead of iterative column-wise and row-wise softmax to achieve a bi-stochastic matrix, we integrate computationally efficient approaches in entropy-regularized optimal transport into our network.


\section{Related Work} 
\label{sec:related_work}

\begin{figure*}[t]
    \centering
    \includegraphics[width=0.8\linewidth]{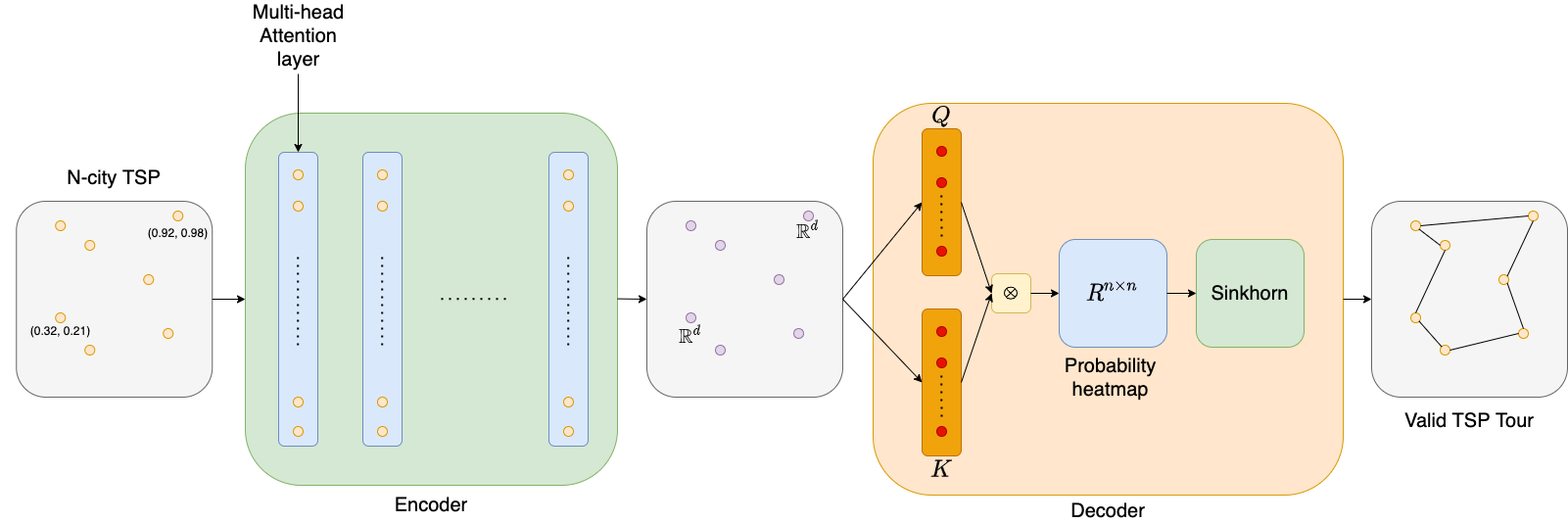}
    \caption{Overview of our approach. We use a standard multi-headed transformer architecture as the encoder. An $n \times n$ heatmap is then produced through vector outer products, which is  converted into a valid distribution by \cite{cuturi2013sinkhorn}'s algorithm. A valid tour is  constructed by decoding. All operations are differentiable and the model is trained end-to-end.}
    \label{fig:model}
\end{figure*}
For brevity, we provide the works most related to our approach. We refer readers to the work in \cite{bresson2021transformer} for a comprehensive summary of traditional algorithms and current deep learning approaches. Recently, there have been many strides in training deep learning models for combinatorial optimisation, specifically the TSP. This can be separated into 2 main approaches: autoregressive approaches and non-autoregressive approaches. 

\subsection{Autoregressive Approaches}

One of the earliest reinforcement learning approaches to the TSP stems from \cite{bello2017neural}, which utilized the Pointer-Network by \cite{vinyals2015pointer} as a model to output a policy which indicates the sequence of cities to be visited by an agent. This policy was trained by policy gradients, and the tours were constructed by either greedy decoding, sampling, or using Active Search. 

\citeauthor{kool2018attention} leveraged the power of attention and transformers introduced in \cite{vaswani2017attention}. They trained the model on randomly generated TSP instances,  utilizing the same policy gradient scheme as \cite{bello2017neural} with a greedy rollout as a baseline. This produced state-of-the-art results with optimality gaps as small as 1.76\% and 0.52\% with greedy search and beam search respectively on TSP50.

More recently, \citeauthor{bresson2021transformer} framed the TSP as a translation problem and utilized a transformer architecture in both the encoder and decoder to construct valid solutions. Likewise, they used a policy gradient approach to train the model and achieved very strong optimality gaps in 0.31\% and 4e-3\% with greedy search and beam search on TSP50.

These solutions tend to decode solutions sequentially in an autoregressive fashion with a learnable decoder, formulating the prediction as maximising the probability of a sequence. Concretely, a $n$-TSP solution can be represented as an ordered list of $n$ cities to visit, where $\text{seq}_{n} = \{ c_1, c_2, ..., c_n \}$. This can be framed as an optimization problem in the form of maximizing the likelihood of the sequence such as

\begin{equation}
    \max_{\text{seq}_n} P^{\text{TSP}}(\text{seq}_n |X) = P^{\text{TSP}}(c_1, ... c_n | X)
\end{equation}

where $X \in \mathbb{R}^{n \times 2}$ represents the set of 2D coordinates for $n$ cities, and $c$ denotes the city index. For autoregressive solutions, $P^{\text{TSP}}$ can be factored using the chain rule of probability, where

\begin{align}
\begin{split}
    P^{\text{TSP}} (c_1, ..., c_n | X) = P(c_1|X) \cdot P(c_2| c_1, X) \cdot \\ ... \cdot P(c_n|c_{n-1}, \dots, c_1, X)
\end{split}
\end{align}

Thus, maximizing the prediction equates to
\begin{equation}
    \max_{c_1, ..., c_n} \Pi^n_{t=1} P(c | c_{t-1}, c_{t-2}, ..., c_1, X)
\label{eqn:pi}
\end{equation}

Essentially, the final joint prediction results from the chain rule of conditional probability, where the next city is predicted based on an embedding of the current partial tour. In both \cite{kool2018attention, bresson2021transformer}, the transformer in the decoder is constantly queried to construct this partial tour embedding to find the next city. In contrast, this conditioning happens naturally in the recurrent mechanism of the Pointer-Network for \cite{bello2017neural}.

\subsection{Nonautoregessive Approaches}
Another approach is to predict the entire adjacency matrix of the TSP and view this matrix as a probability heatmap; each edge denotes the probability of it being in a tour. One of the first attempts at this came from \cite{joshi2019efficient}, which uses a supervised learning approach. Each data sample was first put through an exact solver to generate labels. A graph convolutional neural network is used to encode the cities. The authors reduced the memory usage by only considering the k-Nearest Neighbors for each city to form a graph. They reported an optimality gap of 3.10\% and 0.26\% with greedy search and beam search, respectively.

Following this work, \citeauthor{fu2020generalize} introduced a method to combine small solutions produced by \citeauthor{joshi2019efficient}'s model to form a larger solution. For example, their work uses \citeauthor{joshi2019efficient}'s model to produce heatmaps for a TSP of 20 nodes; they then utilize these heatmaps and combine them via a heuristic to construct a heatmap for a TSP of 50 nodes. This final heatmap then undergoes Monte Carlo Tree Search (MCTS) to generate a large set of possible solutions, and they retrieve the best possible set of solutions from this. Most recently, \citeauthor{kool2021deep} introduced a heuristic dubbed the "Deep Policy Dynamic Programming" to explore the space of possible solutions given \citeauthor{joshi2019efficient}'s model. Essentially, their algorithm provides a form of guided beam search to produce high quality solutions. Combined with multiple efficient implementations, this work produces a set of highly competitive results. Both of these work do not involve training a model, but rather use heuristics for search.

A common procedure in both autoregressive and non-autoregressive solutions is the need for beam search or sampling, as more often than not, local solutions of TSP do not necessarily give the best tour. A key difference in these approaches, is that in the work of \cite{joshi2019efficient}, since edge probabilities are given, the beam search is only used to refine this probability matrix to give a valid tour. Whereas in solutions that rely on sequential decoding such as \cite{kool2018attention, bresson2021transformer}, the transformer embeddings are constantly queried to form a representation of the partial tour. This embedding is then used to find the next city via the attention mechanism. The overall inference times in such models are slow and memory intensive.




\section{Combining RL and Edge Prediction}
\label{sec:method}
Current autoregressive solutions tend to be trained via reinforcement learning which does not rely on labelled data and produce highly competitive solutions. Nonautoregressive solutions, on the contrary, utilizes supervised learning and show their decoding power in speed and scalability, but lack quality. In this section, we take a first stab at marrying the benefits in both of these approaches and show how we leverage optimal transport techniques to make up for the loss in decoding power. Figure \ref{fig:model} depicts the overall structure of our approach, which consists of two main blocks: an encoder to learn latent representations, and a decoder to convert these representations into a valid TSP tour.

\subsection{The Transformer Encoder}
Given an $n$-TSP problem represented by 2D coordinates, $\{x \in X | x \in \mathbb{R}^2, |X| = n \}$, we adopt the standard self-attention transformer layer of \cite{vaswani2017attention} for the encoder model. We utilize residual connections, multi-headed attention, and batch normalisation. Concretely, for a single head attention mechanism, we can view the transformer as
\begin{align}
    H^{l+1} &= \textsc{softmax} \big ( \frac{Q^l K^{l^T}}{\sqrt{d}} V^l \big ) \in \mathbb{R}^{n \times d} \\
    Q^l &= H^l W^l_Q \in \mathbb{R}^{n \times d}, W^l_Q \in \mathbb{R}^{d \times d} \\
    K^l &= H^l W^l_K \in \mathbb{R}^{n \times d}, W^l_K \in \mathbb{R}^{d \times d} \\
    V^l &= H^l W^l_V \in \mathbb{R}^{n \times d}, W^l_V \in \mathbb{R}^{d \times d}
\end{align}
where $l$ denotes a layer of the transformer model, $H$ the hidden representations of dimension $d$, and $H^{l=0} = X$ the set of cities represented by their 2D coordinate points on a plane. At each layer $l$, the representations are projected to $Q, K, V$, the respective query, key, and value representations, using learnable matrices $W_Q, W_K, W_V$. The final output representation is done by computing an attention score via the Softmax of the inner product of $Q$ and $K$. Essentially, each node's representation learns to aggregate information amongst all other nodes.

In order to decide the order of moving from city to city, we produce a probability heatmap, similar to the output in \cite{joshi2019efficient}. We construct an $n \times n$ matrix, $M$, via vector outer products to generate this heatmap. This is done by the following:
\begin{align}
    A &= W_A H \in \mathbb{R}^{n \times d}\\
    B &= W_B H \in \mathbb{R}^{n \times d}\\
    M &= \frac{AB^\top}{\sqrt{d}}\in \mathbb{R}^{n \times n} \\
    P_{\textsc{tanh}} &= \tanh{(M)} * C\in \mathbb{R}^{n \times n}
\end{align}
where $W_A$ and $W_B$ are learnable parameters, $\sqrt{d}$ a scaling factor based on dimensionality of representation, and $H$ denotes the representations given from the final transformer encoder layer. $C$ is a scaling factor to scale the $\tanh$ outputs which helps in gradient flow \cite{bello2017neural}.

Once we have $P_{\textsc{tanh}}$, we explore 2 different types of decoders: (1) Softmax decoder, (2) Sinkhorn decoder. In our work, we view the rows of the $n \times n$ matrix as the "from" city and the columns as the "to" city.

\subsection{The Softmax Decoder}
The Softmax decoder can be viewed as a simple solution to construct a valid distribution. Recall that the heatmap denotes the probability (or log-probability) for an edge to be included in a TSP tour. In order to generate a valid distribution, we can apply the Softmax function row-wise, which represents the probability distribution of moving from one city to another. 

\subsection{The Sinkhorn Decoder} 
While the Softmax function produces a valid distribution and suffices for picking a city, it is short-sighted as it only considers immediate choices. Rather, the decoder should take into account the construction of an entire tour. Additionally, we note that since we have moved away from sequentially encoding the partial tour, we have lost the ability to condition on previous choices. Instead, we opt to introduce useful biases into the model. One observation is that in order to construct a valid TSP tour using a heatmap, the solution of the heatmap has to be a solution to an assignment problem, which can be approximated with the Sinkhorn algorithm. 

In this work, we adopt \citeauthor{cuturi2013sinkhorn}'s approach in calculating the optimal transport plan. For two probability distributions $r$ and $c$ residing in a simplex $\Sigma_d := \{ x \in \mathbb{R}^d_+ : x^\top \mathbf{1}_d = 1 \}$, where $\mathbf{1}_d$ is a dimensional vector of ones with size $d$, the transport polytope of this problem lies in the polyhedral set of $d \times d$ matrices, and is given by
\begin{equation}
    U(r,c) := \{ T \in \mathbb{R}^{d \times d}_{+} | T\mathbf{1}_d = r, T^\top \mathbf{1}_d = c \}
\end{equation}

Hence, $U(r,c)$ contains all nonnegative $d \times d$ matrices, where $r$ and $c$ are the row and column sums. Note that since $r$ and $c$ are probability distributions, their row and column sums thus amount to 1 respectively. Therefore, $U(r,c)$ contains the joint probability distributions for two multinomial random variables across values $\{1,\dots,d\}$. Suppose a matrix $T \in U(r,c)$ denotes a joint distribution across multinomial random variables $(X,Y)$ such that $p(X=i, Y=j) = t_{ij}$ is an entry of the matrix $T$ at row $i$ and column $j$, the optimal transport problem is defined as:
\begin{equation}
    d_M (r,c) := \min_{T \in U(r,c)} \langle T,M \rangle
\end{equation}

where $M$ is a cost matrix given by the problem. In the most extreme case, $T$ can be a binary-only matrix, where each entry is either 1 or 0. Given the constraints of the problem where the row and column marginals of $T$ must sum to 1, this extreme case becomes a solution of the assignment problem. 
For a given optimal transport plan, the entropy can be calculated as
\begin{equation}
    h(T) = - \sum^d_{i,j=1} t_{ij} \log t_{ij}
\end{equation}
Concretely, the optimization problem is now given by:
\begin{equation}
    T^\lambda = \argmin_{T \in U(r,c)} \langle T, M \rangle - \frac{1}{\lambda} h(T), \text{ for } \lambda > 0
    \label{eqn:sink_cuturi}
\end{equation}

\begin{figure}[t]
    \centering
    \includegraphics[width=0.8\linewidth]{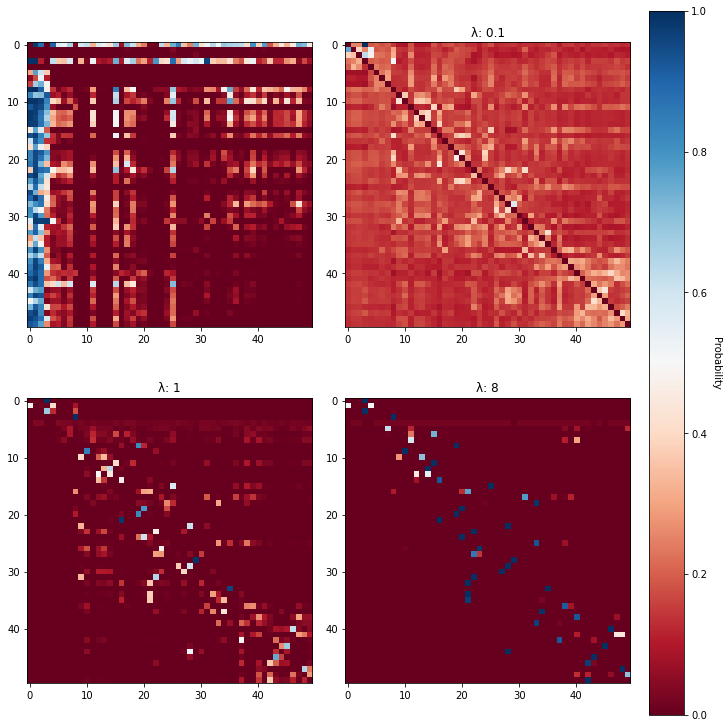}
    \caption{\citeauthor{cuturi2013sinkhorn}'s algorithm applied to an output from our model for a TSP50 problem. Best viewed in color. Top left depicts the original probabilities, while the respective values of $\lambda$ are indicated accordingly. Colors range from red (close to 0) to blue (close to 1), and diagonals have been masked away as it is not a valid choice. Without any enforcement of assignment constraints, the network's predictions become concentrated, and following these probabilities does not result in a valid tour. Enforcing these constraints results in the distribution spreading out across the cities. Increasing $\lambda$ results in less entropic regularization and the optimal transport map becomes closer to the assignment solution, with strong probability at a single point and others coming close to 0. With more regularization, the problem moves away from the corner of the simplex and the distribution becomes more even and uniform.}
    \label{fig:cut_algo_img}
\end{figure}

As described, the extreme case of optimization might result in a valid assignment solution which is discrete in nature. Such solutions are poor for deep learning models, as we require smooth landscapes for optimization and gradient backpropagation. \citeauthor{cuturi2013sinkhorn}'s introduction of entropic regularization shifts the solution away from the corner of the simplex, providing a control on its sparsity, and more importantly, a valid distribution. The resulting method results in a bi-stochastic matrix, and it also leverages vector-matrix multiplications for fast calculations via the GPU. Hence, we view this algorithm as a suitable candidate to integrate with a deep learning network to produce a valid TSP solution. For our approach, we can view the final output of $P_{\textsc{tanh}}$ as a cost matrix, and by passing it through the Sinkhorn algorithm, we enforce the bi-stochastic constraints on the output. Note that these constraints are enforced on the entire matrix, and hence a global condition to satisfy when constructing tours. Since all operations are differentiable, the network can be trained end-to-end to learn representations that respect these constraints. Algorithm \ref{alg:cap} showcases our implementation of \citeauthor{cuturi2013sinkhorn}'s in the context of our model, and Figure \ref{fig:cut_algo_img} provides some intuition on the algorithm.

\begin{table}[t]
\small
\begin{center}
\begin{tabular}{l|c|c|c|c}
\multicolumn{1}{l}{} & \multicolumn{1}{l}{} & \multicolumn{1}{l}{}& \multicolumn{2}{c}{Time}\\
\multicolumn{1}{l|}{Model}                        & \multicolumn{1}{c|}{\# Params} & \multicolumn{1}{c|}{GPU (Gb)} & \multicolumn{1}{c|}{Epoch}   & Total   \\ \hline
\multicolumn{1}{l|}{\textbf{AM}} & \multicolumn{1}{c|}{708, 352}  & \multicolumn{1}{c|}{4.85}     & \multicolumn{1}{c|}{7m 2s}   & 11h 42m \\
\multicolumn{1}{l|}{\textbf{TM}} & \multicolumn{1}{c|}{1,009,152} & \multicolumn{1}{c|}{7.94}     & \multicolumn{1}{c|}{13m 30s} & 22h 30m \\
\multicolumn{1}{l|}{Softmax}                      & \multicolumn{1}{c|}{1,091,200} & \multicolumn{1}{c|}{4.33}     & \multicolumn{1}{c|}{4m 7s}   & 6h 51m  \\
\multicolumn{1}{l|}{Sinkorn}                      & \multicolumn{1}{c|}{1,091,200} & \multicolumn{1}{c|}{4.35}     & \multicolumn{1}{c|}{4m 15 s} & 7h 5m  
\end{tabular}%
\end{center}
\caption{Illustration of model capacity, GPU memory usage, average time for 1 epoch (epoch time), and total training time for 100 epochs, for the TSP50 problem. \textbf{AM} refers to the attention model by \cite{kool2018attention}, and \textbf{TM} refers to the transformer model by \cite{bresson2021transformer}. Our model is at least 10\% lighter on memory usage, and is at least 40\% faster.}
\label{tbl:params}
\end{table}

\begin{algorithm}
\caption{Sinkhorn Matrix Calculation}\label{alg:cap}
\begin{algorithmic}
\Require $P_{\textsc{tanh}}, \lambda$, \text{number of iterations I}
\State $K = \exp{(-\lambda * P_{\textsc{tanh}}}) \in \mathbb{R}^{n \times n}$
\State $u \gets \frac{\mathbf{1}}{n} \in \mathbb{R}^{n}$
\State $v \gets \frac{\mathbf{1}}{n} \in \mathbb{R}^{n}$
\For{i in 1:I}
    \State $\Tilde{K} = K^\top u \in \mathbb{R}^{n}$
    \State $v = 1 ./ \Tilde{K} \in \mathbb{R}^{n}$
    \State $u = 1 ./ (Kv) \in \mathbb{R}^{n}$
\EndFor
\State $P = \text{diag}(u)  K  \text{diag}(v) \in \mathbb{R}^{n \times n}$
\State $P_{\textsc{logits}} = \log{(P)} \in \mathbb{R}^{n \times n}$ 
\\
\State where $./$ denotes the element-wise division.
\end{algorithmic}
\end{algorithm}


\begin{table*}[t]
\begin{center}
\small 
\begin{tabular}{lcc|c|c|ccc}
 &
  \multicolumn{1}{l}{} &
  \multicolumn{3}{c|}{\textbf{TSP50}} &
  \multicolumn{3}{c}{\textbf{TSP100}} \\
\multicolumn{1}{l|}{Model} &
  \multicolumn{1}{c|}{Search Type} &
  Tour Length &
  Opt. Gap &
  Time &
  \multicolumn{1}{c|}{Tour Length} &
  \multicolumn{1}{c|}{Opt. Gap} &
  Time \\ \hline
\multicolumn{1}{l|}{Concorde} &
  \multicolumn{1}{c|}{-} &
  5.689* &
  0.00\% &
  2m* &
  \multicolumn{1}{c|}{7.765*} &
  \multicolumn{1}{c|}{-} &
  3m* \\
\multicolumn{1}{l|}{\textbf{AM}} &
  \multicolumn{1}{c|}{Greedy} &
  5.790 &
  1.77\% &
  1.3s &
  \multicolumn{1}{c|}{8.102} &
  \multicolumn{1}{c|}{4.34\%} &
  2.1s \\
\multicolumn{1}{l|}{\textbf{AM}} &
  \multicolumn{1}{c|}{Sampling, 1280} &
  5.723 &
  0.59\% &
  11m33s &
  \multicolumn{1}{c|}{7.944} &
  \multicolumn{1}{c|}{2.30\%} &
  29m 36s \\
\multicolumn{1}{l|}{\textbf{TM}} &
  \multicolumn{1}{c|}{Greedy} &
  \underline{5.753} &
  \underline{1.12\%} &
  10.7s &
  \multicolumn{1}{c|}{\underline{8.028}} &
  \multicolumn{1}{c|}{\underline{3.39\%}} &
  13.8s \\
\multicolumn{1}{l|}{\textbf{TM}} &
  \multicolumn{1}{c|}{Beam search, 2500} &
  \textbf{5.696} &
  \textbf{0.12\%} &
  21m 26s &
  \multicolumn{1}{c|}{\textbf{7.871}} &
  \multicolumn{1}{c|}{\textbf{1.37\%}} &
  1h 30m
  \\
\multicolumn{1}{l|}{Softmax} &
  \multicolumn{1}{c|}{Greedy} &
  5.841 &
  2.68\% &
  0.24s &
  \multicolumn{1}{c|}{8.280} &
  \multicolumn{1}{c|}{6.76\%} &
  0.85s \\
\multicolumn{1}{l|}{Softmax} &
  \multicolumn{1}{c|}{Beam search, 2500} &
  5.757 &
  1.38\% &
  2m 17s &
  \multicolumn{1}{c|}{8.143} &
  \multicolumn{1}{c|}{4.87\%} &
  \textbf{9m 26s} \\
\multicolumn{1}{l|}{Sinkhorn} &
  \multicolumn{1}{c|}{Greedy} &
  5.782 &
  1.62\% &
  \underline{0.24s} &
  \multicolumn{1}{c|}{8.110} &
  \multicolumn{1}{c|}{4.44\%} &
  \underline{0.85s} \\
\multicolumn{1}{l|}{Sinkhorn} &
  \multicolumn{1}{c|}{Beam search, 2500} &
  5.719 &
  0.53\% &
  \textbf{2m 14s} &
  \multicolumn{1}{c|}{8.000} &
  \multicolumn{1}{c|}{3.03\%} &
  9m 38s
\end{tabular}
\end{center}
\caption{Optimality gaps and average tour lengths for various models on TSP50 and TSP100 on 10,000 samples. Entries with an asterisk(*) denotes that the result was taken from other papers. Bold entries denote best performance for sampling search while underlined entries refer to greedy search. \textbf{AM} refers to the attention model by \cite{kool2018attention}, and \textbf{TM} refers to the transformer model by \cite{bresson2021transformer}. Our solution shows significant speed increases while still being able to achieve reasonable performance. Additionally, the benefits of Sinkhorn bias is clear as the gap reduces while showing no slow down in inference.}
\label{tbl:opt_gap}
\end{table*}

\subsection{Decoding Process}
Once we retrieve $P_{\textsc{logits}}$, we can decode a valid tour a single step at a time. While \citeauthor{bresson2021transformer} utilizes a learnable token to indicate the starting city, we found that we can achieve similar performance by fixing the first city always to be the top right-hand most city (closest to coordinate (1,1)). To achieve this, we pre-sort the data by the sum of their coordinates and order them accordingly. This has minimal computational cost and removes the need for any extra compute in the GPU for the token. 

During training, we categorically sample actions to take according to their log-probabilities, so that the agent is able to explore some solutions instead of always taking a greedy action. At each step of decoding, we apply a masking procedure to remove cities that have been visited and self-loops, followed by re-normalizing the distributions to ensure a valid distribution is always produced. 

The entire encoding and decoding process can be viewed as a parameterized policy, $\pi_\theta$, which first encodes all cities to a latent representation, followed by producing a trajectory of cities to visit. As discussed previously, autoregressive approaches condition on previous predictions displayed in equation \ref{eqn:pi} by encoding partial tours. In our method, we produce this by decoding the $n \times n$ matrix via the re-normalizing method discussed above. Essentially, we produce all probabilities first, and simply re-normalize them as the agent takes actions. Notably, since we do not have to encode the partial tours for conditioning such as in methods by \cite{kool2018attention} and \cite{bresson2021transformer}, our approach becomes much faster as the computational graph is greatly reduced.

\subsection{Learning Algorithm}
In this work, we adopt the REINFORCE algorithm \cite{williams1992simple}. Our model produces a heatmap in the form of $P_{\textsc{logits}}$, where each entry $P_{ij}$ represents the log-probability of moving from city $i$ to city $j$. As mentioned, the model produces a trajectory during the decoding process. We accumulate these log-probabilities and calculate the policy gradient as:
\begin{equation}
    \nabla \mathcal{L} (\theta | x) = \mathbb{E}_{p_\theta (\pi | x)} [(L(\pi) - b(x)) \nabla \log p_\theta (\pi|x)]
    \label{eqn:rl_algo}
\end{equation}

\noindent where $L(\pi)$ denotes the length of a tour following our policy $\pi$, $p_{\theta}$ denotes the predicted probability of moving to a city by following the policy $\pi_{\theta}$, and $b(x)$ denotes the length of a tour following a greedy roll-out of a baseline model, similar to the work in \cite{kool2018attention, bresson2021transformer}. Additionally, we only update the parameters of this baseline model if the mean length of the tour on a generated validation set is less than a threshold, similar to that of \cite{kool2018attention, bresson2021transformer}.

\subsection{Inference}

Once the model is trained, we look at two approaches to produce valid trajectories: greedy search and beam search. Similar to the work done by \cite{joshi2019efficient, kool2018attention, bresson2021transformer}, we either perform a greedy decode or a beam search at test time to explore the space of solutions, reporting the shortest length tour found.



\section{Experiments \& Results} \label{sec:exp}

\subsection{Experiment Setup}

\textbf{Resources: } All experiments are run on a Nvidia DGX Workstation machine with 4 A100 GPUs. All experiments are run on a single GPU only.

\textbf{Dataset: } Since this is a reinforcement learning problem, we do not use any labelled data. Instead, at each epoch, a batch of data is generated for training. At the end of an epoch, a set of 10,000 TSP instances are generated to evaluate the performance for updating the baseline. Finally, a fixed set of pre-generated 1,000 TSP instances are used to calculate the optimality gaps to validate performance.

In order to judge the performance of our model fairly, we utilize the same number of training epochs as \cite{kool2018attention}, 100 epochs, as it dictates the total amount of data used. Essentially, we train on 2500 generated batches, each of size 512, resulting in 1,280,000 generated TSPs.

\textbf{Models: }
We compare both the Softmax and Sinkhorn approaches, as well as benchmark our results to other state-of-the-art reinforcement learning models \citep{kool2018attention, bresson2021transformer}. For the Sinkhorn approach, we utilize one iteration of the algorithm with a regularization $\lambda = 2$. Table \ref{tbl:params} displays the various models, total parameter budget, GPU usage, and runtimes for training. For all experiments, we use the ADAM optimizer \cite{kingma2014adam} with gradient clipping.


\begin{table}
\small
\centering
\begin{tabular}{c|c|c|c|c|c}
$\lambda$ & T & G Score & G Gap & BS Score & BS Gap \\ \hline
2.0    & 1             & 5.782        & 1.62\%     & 5.719             & 0.53\%          \\
3.3    & 1             & 5.795        & 1.87\%     & 5.726             & 0.65\%          \\
2.0    & 3             & 5.786        & 1.69\%     & 5.719             & 0.53\%          \\
1.0    & 1             & 5.841        & 2.67\%     & 5.761             & 1.27\%          \\
1.0    & 3             & 5.854        & 2.90\%     & 5.771             & 1.43\%         
\end{tabular}%
\caption{Comparison of regularization and number of iterations for \citeauthor{cuturi2013sinkhorn}'s algorithm. G and BS refer to greedy and beam search decoding, respectively. Scores indicate the tour length while gap refers to the optimality gap compared to Concorde. Smaller $\lambda$ indicates more entropic regularization and vice versa. Results are on TSP50.}
\label{tbl:reg_iter}
\end{table}
\textbf{Evaluation Metrics: } We evaluate the model based on a pre-generated set of 10,000 TSP instances. We compute the optimality gap of the solution as follows:
\begin{equation}
    \textsc{optimality gap} = \frac{\frac{1}{N} \sum^{N}_{i=1} L(\pi_{\theta})^{(i)}}{\frac{1}{N} \sum^{N}_{i=1} L^{(i)}_{T}}
\end{equation}

where $L(\pi_{\theta})^{(i)}$ denotes the tour length based on the output of our model on one instance $i$, and $L^{(i)}_T$ the tour length based on the output of a Concorde solver on instance $i$. For our experiments, $N$ refers to the 10,000 TSP instances.

We compare all approaches based on the length of the tour, optimality gap, memory usage, and wall-clock runtimes. We also note that both recent approaches in \cite{fu2020generalize} and \cite{kool2021deep} that leverage the supervised learning model in \cite{joshi2019efficient} outperform our models in accuracy but not speed.

\subsection{Results}

\textbf{Computational Complexity: } Table \ref{tbl:params} describes the overall complexity of the models and runtimes. Notably, removing a learnable decoder results in training a model that is almost twice as fast and uses at least 10\% less GPU memory per pass. Note that we also have more parameters but still less overall GPU usage compared to \cite{kool2018attention}. Additionally, the inclusion of \citeauthor{cuturi2013sinkhorn}'s algorithm has minimal impact in both GPU usage and training time.

\textbf{Overall Performance: } Table \ref{tbl:opt_gap} showcases the various model performances across 10,000 TSP instances. We rerun all models again on our machine to get updated results for a fair comparison. Evidently, moving away from a learnable decoder reduces the model's predictive power; the self-attention transformer is highly capable of encoding partial tours that improve the overall performance. However, we also see that the inference times in our approach has improved tremendously. At the same time, we can achieve better performance than the previous state-of-the-art in \cite{kool2018attention}. Also, since we can run beam search much faster, this suggests that we can increase the total number of beams, improving the search space and solution quality. With the addition of the optimal transport bias into the decoder, we see a 3-fold improvement in optimality gap, with minimal impact on inference time on TSP50. Likewise, experiments on TSP100 produce similar observations; the optimal transport bias is beneficial to the model and is light. 

\textbf{Impact of Regularization and Iterations: } Table \ref{tbl:reg_iter} displays various models we trained based on different values of $\lambda$ and the number of iterations. It would appear that the regularization value is essential, too much regularization degrades performance. At the same time, the number of iterations has less impact when there is sufficient regularization.

\section{Discussion \& Future Work}

\textbf{Scaling to larger TSPs: } We note that works such as \cite{fu2020generalize, kool2021deep} are able to predict on larger TSPs. However, a key difference between our work and these is that these works rely on a pre-trained supervised learning model in \cite{joshi2019efficient}, whereas ours works solely on reinforcement learning. Since both our model and \citeauthor{joshi2019efficient} aim to produce edge probabilities, we plan to investigate if we are able to learn strong edge probabilities for larger instances since our approach is faster and less compute intensive than other reinforcement learning approaches.


\textbf{Improving the search strategy: } Another benefit of our approach is that since we only compute the edge probabilities, we can use a variety of search strategies. Both \cite{fu2020generalize, kool2021deep} use various search strategies in MCTS and DPDP. We plan to see how our model predicts with such search strategies as well.

\textbf{Including search during training: } For most reinforcement learning approaches, greedy search or categorical sampling is used during training to retrieve the trajectories. However, this is rather inefficient early in the learning phase as the model has no idea what is a good solution yet. Since our model is light, we can potentially include search procedures such as beam search during training instead to retrieve good quality solutions to learn from.


\section{Conclusion}

In this work, we have shown that moving away from a learnable decoder loses predictive power but provides clear benefits in computational complexity and runtimes. In order to make up for this loss, we investigate the use of optimal transport algorithms by incorporating them as a differentiable layer. This addition provides minimal impact on the complexity and runtimes, yet it has clear advantages in performance. These contributions opens the door for investigating how various search strategies can now be used to retrieve strong solutions for the TSP, as well as being able to train directly on larger TSPs.




\section*{Acknowledgments}
This work was funded by the Grab-NUS AI Lab, a joint collaboration between GrabTaxi Holdings Pte. Ltd. and National University of Singapore, and the Industrial Postgraduate Program (Grant: S18-1198-IPP-II) funded by the Economic Development Board of Singapore. Xavier Bresson is supported by NRF Fellowship NRFF2017-10 and NUS-R-252-000-B97-133.

 \typeout{}
\bibliography{refs}

\end{document}